\documentclass{article}

\usepackage{arxiv}

\usepackage[utf8]{inputenc} 
\usepackage[T1]{fontenc}    
\usepackage{hyperref}       
\usepackage{url}            
\usepackage{booktabs}       
\usepackage{amsfonts}       
\usepackage{nicefrac}       
\usepackage{microtype}      
\usepackage{lipsum}
\usepackage{graphicx}
\usepackage{authblk}
\usepackage{float}
\graphicspath{ {./images/} }

\usepackage{amsmath,amssymb}
\usepackage{comment}
\usepackage{mathrsfs}
\usepackage{graphicx,nicefrac}
\usepackage{float}

\usepackage{comment}
\usepackage{scalerel}
\usepackage[overload]{empheq}
\usepackage{textgreek}

\usepackage{algorithm}
\usepackage{algpseudocode}
\usepackage{enumitem}

\usepackage{mathrsfs}
\usepackage{graphicx,nicefrac}

\usepackage[utf8]{inputenc}
\usepackage[english]{babel}

\usepackage[margin=1cm]{caption}
\usepackage{comment}
\usepackage{scalerel}
\usepackage[overload]{empheq}

\usepackage{float}

\usepackage{authblk}
\title{Increasing a microscope's effective field of view via overlapped imaging and machine learning}

\author[1]{Xing Yao}
\author[1]{Vinayak Pathak}
\author[2]{Haoran Xi}
\author[1]{Amey Chaware}
\author[3]{Colin Cooke}
\author[1]{Kanghyun Kim}
\author[1]{Shiqi Xu}
\author[1]{Yuting Li}
\author[1,4]{Timothy Dunn}
\author[1]{Pavan Chandra Konda}
\author[1]{Kevin C. Zhou}
\author[1,3,*]{Roarke Horstmeyer}

\affil[1]{Department of Biomedical Engineering, Duke University, Durham, NC, USA, 27708}
\affil[2]{Department of Computer Science, Duke University, Durham, NC, USA, 27708}
\affil[3]{Department of Electrical and Computer Engineering, Duke University, Durham, NC, USA, 27708}
\affil[4]{Department of Neurosurgery, Duke University, Durham, NC, USA, 27708}
\affil[*]{roarke.w.horstmeyer@duke.edu}

\begin{document}
\maketitle
\begin{abstract}
This work demonstrates a multi-lens microscopic imaging system that overlaps multiple independent fields of view on a single sensor for high-efficiency automated specimen analysis. Automatic detection, classification and counting of various morphological features of interest is now a crucial component of both biomedical research and disease diagnosis. While convolutional neural networks (CNNs) have dramatically improved the accuracy of counting cells and sub-cellular features from acquired digital image data, the overall throughput is still typically hindered by the limited space-bandwidth product (SBP) of conventional microscopes. Here, we show both in simulation and experiment that overlapped imaging and co-designed analysis software can achieve accurate detection of diagnostically-relevant features for several applications, including counting of white blood cells and the malaria parasite, leading to multi-fold increase in detection and processing throughput with minimal reduction in accuracy.
\end{abstract}

\section*{Introduction}

\label{sec:intro}
Automatic analysis of cells, microorgansims, or other subcellular features within microscope images is essential for a wide range of biomedical and diagnostic applications. Over the past several years, the application of convolutional neural networks (CNNs) has dramatically improved the performance of different counting tasks in various scenarios, including for cancer and tumor diagnosis\cite{9,10}, infectious disease detection\cite{47,11}, and computerized automation of complete blood cell (CBC) counting\cite{46}. However, the overall efficiency of the automated microscope-based cell counting is still constrained by the limited field of view (FOV) of conventional microscopes, especially for tasks which require high-resolution images \cite{13}. This limitation is compounded by the fact that for many applications, the diagnostically-relevant features of interest is highly sparse. For example, to diagnose infection with the malaria parasite (\textit{Plasmodium falciparum}), a trained expert often needs to examine a blood smear with a 100$\times$ microscope objective to identify the parasite\cite{12} across several hundred fields of view, with the aim of visually identifying just one or a few parasites \cite{13}, which can lead to a serious bottleneck within the diagnostic pipeline. 

A core reason why we cannot simultaneously obtain high spatial resolutions over large FOVs stems from lens aberrations. The higher the desired spatial resolution, the more difficult it is for lens designers to correct for aberrations at off-axis positions. As a result, most standard microscopes are limited to a space-bandwidth-product (SBP) of 10s of megapixels \cite{1}. Current solutions to the limited SBP problem include whole slide imaging (WSI) scanners \cite{53,54}, Fourier ptychographic microscopy (FPM)\cite{6,konda2020fourier,zheng2021concept}, and multi-aperture systems \cite{brady2012multiscale,fan2019video}. However, WSI requires high-precision mechanically scanning of the sample or the imaging system \cite{bueno2014automated}, which in turn may require repeated focus adjusting, making the entire process time consuming and expensive \cite{evans2018us}. 

In this work, we propose a new solution that overlaps multiple microscope images from different FOVs onto a single image sensor. After acquiring a single snapshot, a machine learning algorithm then extracts relevant information over the larger effective FOV for cell counting tasks. 
Our method takes advantage of two qualities of many cell counting tasks. First, the target itself is often sparse and thus has low probability of overlapping with each other in the composite image. Second, the positions of the cells are not important, and thus, unlike previous overlap-based multiplexed imaging approaches \cite{15,16,17}, we do not need to reconstruct the extended FOV, but rather the CNN can be trained and deployed on image patches \cite{44}. Thus, in forming superimposed images, our approach is efficient not only in the data capture, but also in the computational analysis. We demonstrate our approach in experiment on the specific clinical tasks of white blood cell (WBC) \cite{50,49} and malaria parasite counting \cite{51}.

\begin{figure}[!t]
\centering
\includegraphics[width=13cm]{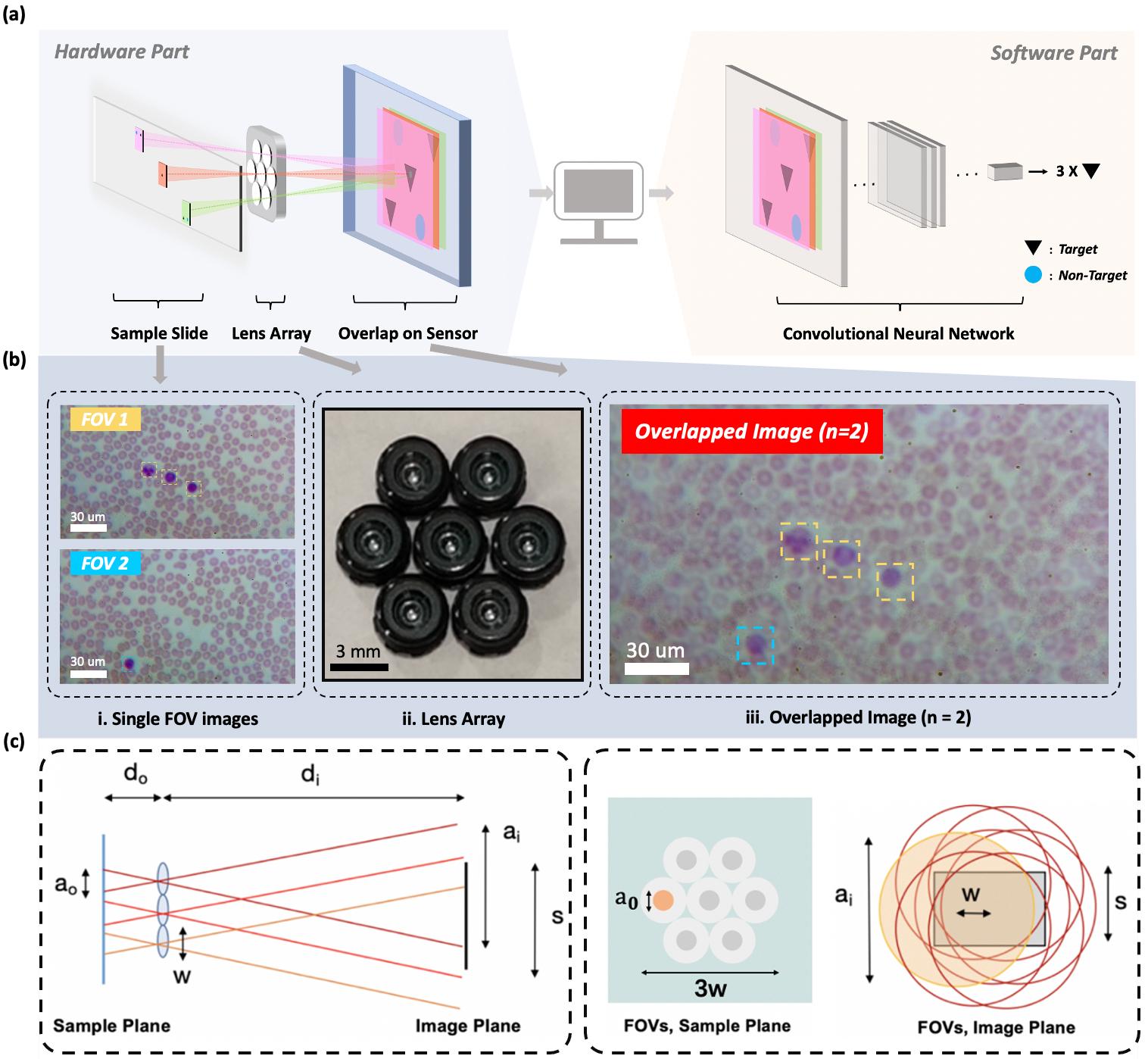}
\caption{(a) Pipeline of overlapped microscope imaging. Multiple independent sample FOVs are illuminated by LEDs and then imaged onto a common sensor through a multi-lens array to form an overlapped image. A CNN-based post processing framework is then employed to identify target features or objects within the overlapped image. (b) i. 2 single-FOV images (FOV1, FOV2) of a hematological sample, ii. the lens array, and iii. the overlapped image ($n$ = 2) for WBC counting task. (c) Overlapped imaging setup. Left: imaging geometry with sample imaged by three lenses and overlapped on a single image sensor. Right: FOVs of 7 lenses marked as gray circles at the sample plane, where FOVs of diameter $a_o$ are separated by lens pitch $w$ and do not overlap. At the image plane, FOVs denoted by red circles of diameter $a_i$ have significant overlap. Marked variables are listed in Table 1.}
\label{pipline}
\end{figure}

\section{Proposed method}
\label{sec:headings}

\subsection{Principle and design of overlapped imaging}
Fig. \ref{pipline} presents an overview of the general principle of overlapped imaging for rapid classification and counting. Instead of relying on a single objective lens, our goal here is to use an array of sub-lenses, each of which images a unique, independent FOV onto a common sensor. With $n$ sub-lenses, our approach can in principle capture light from an $n$ times larger FOV as compared with a standard microscope, albeit from disjoint slide areas (i.e., FOVs that are not necessarily directly adjacent). While the individual sub-images overlap and thus reduce image contrast, experimental results show this approach is still effective for tasks where the goal is to search for sparsely distributed features across a relatively uniform or repetitive background, such as detecting WBCs or malaria parasites from blood smears.

For most of our experiments, we used $n=7$ sub-lenses (Fig. \ref{pipline}b(ii)) to image up to seven unique FOVs, as this number of lenses leads to an efficient hexagonal packing geometry and offers a good balance between the increase in effective FOV and the proportional decrease in dynamic range. In our imaging configuration, all lenses are placed parallel to the image sensor (Sony IMX477R) to ensure all the FOVs are approximately in focus at the same plane. The sub-lens size, spacing, object distance, and image distance (detailed in Table 1) were chosen to ensure that the individual image from every sub-lens covers the majority of the sensor. While WBC and malaria parasite detection tasks are traditionally implemented with microscopes with more than 40$\times$ magnifications, here we used slightly lower resolution (25$\times$), based upon findings in our prior work \cite{24}. 

Specifically, we set the working distance to the microscope slide, $d_o$, to 1.2 mm and the image distance, $d_i$, to 30 mm, creating $M=25\times$ magnification sub-images. The diameter of each sub-FOV at the object plane, $a_o$, and its associated diameter at the image plane, $a_i$, are set by the selected lens and obey the relationship $a_i = M × a_o$. In our experiments, $a_o$ = 0.84 mm and $a_i$ = 21 mm. The distance between each lens within our lens array, $w = 3.7$ mm, defines the distance between the center of the FOV of each sub-lens at the sample plane. Each sub-FOV of diameter $a_o$ = 0.84 mm was thus separated from adjacent sub-FOVs by approximately $w = 3.7$ mm as well (Fig. \ref{pipline}b(i,iii)). Using $n$ = 7 lenses in total leads to a total array diameter of approximately $3w = 11$ mm, which approximately matches the width of a typical blood smear slide (typically around 1.5 cm $\times$ 2 cm). Pinhole illumination arrays are also used to prevent cross-talk of light from individual FOVs. A complete list of the parameters used in our initial experiments is presented in Table 1.

\begin{table}[!t]
\footnotesize
\label{table::parameter}
\caption{Table of parameters, overlapped microscope}
\centering
\begin{tabular}{c|c|c}   
 Parameter & Variable & Value \\  
\hline 
  Number of lenses  & $n$  & 7 \\
  Object distance  &$ d_{o}$  & 1.2 mm \\
  Image distance  & $ d_{i}$  & 30 mm \\
  Inter-lens spacing  & $w$  & 3.7 mm \\
  Magnification  & $M$  & 25$\times$ \\
  Resolution(full pitch)  & $d_x$  & 2 $\mu$m \\
  Numerical aperture  & NA & 0.25 \\
  Sub-FOV diameter, object  & $ a_{o}$ & 0.84\\
  Sub-FOV diameter, image  & $ a_{i}$ & 21 mm \\
  Total lens array width & $3w$ & 11.1 mm \\
  Sensor dimension (width) & $s$ & 6.287 mm\\
  Pixel size & $\partial p$ & 1.55$\mu$m \\
  Number of pixels & $p$ & 12.3 MP\\
\end{tabular}
\end{table}

Although overlapped imaging increases the effective FOV of our microscope, the general approach suffers from two disadvantages: first, the dynamic range of each sub-image, $D$, decreases with an increase in the image overlap number, $n$. Image sensors exhibit a limited total dynamic range, $D_t$. Typically, $D_t=256$ grayscale values per pixel for an 8-bit sensor, so to avoid sensor saturation, these values must be divided amongst all sub-images. On average, we can expect the sub-image dynamic range to be $D = D_t/n$. For example, with $n$ = 7 overlapped images, an 8-bit sensor can only dedicate 36 - 37 grayscale values to each sub-image. For tasks where grayscale variations are important for accurate classification, it is beneficial to perform overlapped imaging with either a high-dynamic-range sensor or a high-dynamic-range capture strategy on a standard sensor. Second, the signal-to-noise ratio (SNR) of each sub-image decreases with an increase in the image overlap number, $n$. Assuming that the image sensor can capture $m$ photons before saturation when detecting just a single non-overlapped image ($n$ = 1), there are on average $m/n$ photons per sub-image. Assuming shot noise as the dominant noise source, our SNR per sub-image scales with the square root of the number of photons per sub-image, 1/ $\sqrt{n}$.

As we will see, these disadvantages, nonetheless, may not dramatically impact the accuracy of current deep learning-based classification for a wide selection of tasks. Images often contain a high amount of redundancy, especially when the end goal is a global classification decision. To verify our idea, we show how WBCs and malaria parasites can still be accurately detected from overlapped images with extended FOVs in both simulation data with a realistic noise model and experimental data from our prototype microscope.

\subsection{Deep learning-based cell detection}
In this work, we adopted a 10-layer VGG-like\cite{36} framework with He-normal initialization\cite{he2015delving} and leaky ReLU activation to detect target objects from acquired overlapped image data. We use five sets of $3\times3$ convolutional filters and a single fully connected layer, with each set containing two 2D convolutions with the second convolution having a nonunary stride to reduce the spatial dimension of the tensor.

In a number of learning-based tasks, annotation of regions of interest (ROIs), such as bounding boxes and segmentation masks, are desired. To avoid the relatively tedious process of segmentation mask annotation and to ensure our model remained as widely applicable to different input data types as possible, regardless of the requirement of special annotation, we adopted a binary-classification framework for image post-analysis. During training, we utilized image patches as input. During testing and experimental use across full-FOV image data, we then generated classification heatmaps via a standard sliding window approach\cite{74,75,76}, which is widely used for spatial density prediction. 
Our networks' resulting heatmaps include the probability of the presence of target objects at each pixel over the whole FOV of the various overlapped images, from which various statistics, including counting, are subsequently derived.  

\begin{figure}[!t]
\centering
\includegraphics[width=13cm]{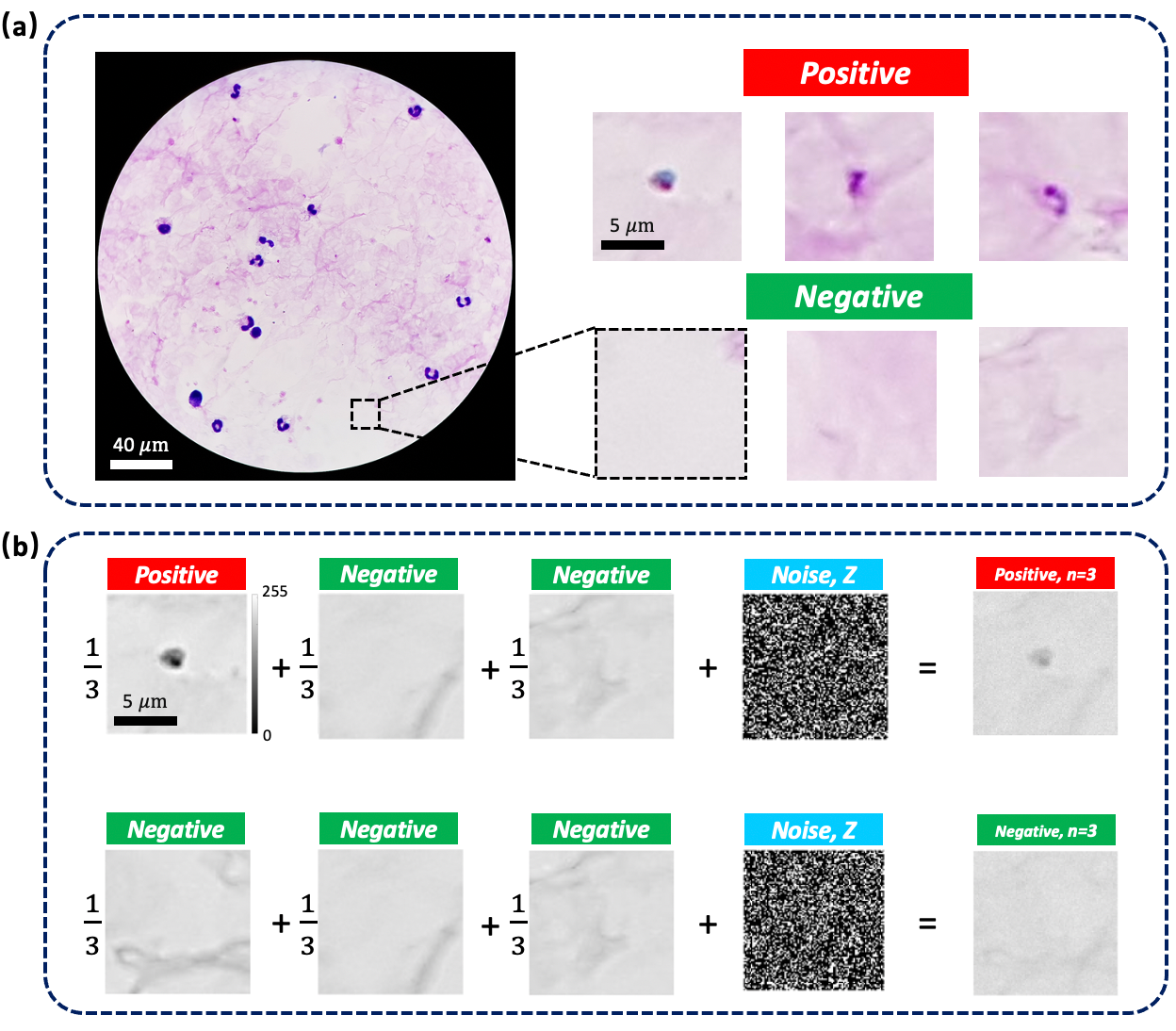}
\caption{Simulation of overlapped imaging for malaria parasite identification. (a) High-resolution, large-FOV images of malaria-infected thick blood smears (from \textbf{SimuData}) were cropped into $96\times96$ patches, and labeled as positive or negative based on the presence of the malaria parasite. (b) We generate digitally overlapped images by averaging patches and adding pixel-adaptive Gaussian noise to compensate for an artificially inflated SNR. Malaria-positive overlapped patches consisted of 1 malaria-infected patch and $n-1$ negative patches. }
\label{simu_demo}
\end{figure}

\section{Simulation procedures and datasets}
\subsection{Simulation of overlapped images with correct noise statistics}
We first demonstrated our method in simulation by applying it to digitally overlapping images (Fig. \ref{simu_demo}). For a given number of lenses $n$, we created synthetic overlapped imaging datasets by digitally averaging $n$ images, yielding  $X_\mathit{avg}$. To counteract the resultant artifactual $\sqrt{n}$ improvement in SNR, as expected from averaging $n$ independent measurements, we added Gaussian noise $Z$, whose standard deviation was scaled based on the averaged value at each pixel location. Finally, the resultant image was quantized into 8 bits, simulating sensor digitization. See Appendix for noise model details. 


For each cell-counting task, we dynamically overlapped small image patches across the available FOVs. The overlapped images were assigned labels based on the presence of target objects in at least one of the sub-FOV images. Fig. \ref{simu_demo}b illustrates this process for the $n$ = 3 case. Our synthetic approach allowed us to easily vary the number of overlapped images. We capitalized on this degree of freedom to characterize the performance of our system as a function of number of overlapping images for each task. This method of synthetically creating overlapped images also serves to augment our data for CNN training -- by varying which regions are overlapped, the number of overall unique overlapped images increases combinatorially with $n$. 


\subsection{Datasets for cell-counting tasks}
\label{datasets}
\subsubsection{SimuData} We first conducted a simulation experiment to study the effectiveness of our method to automatically detect malaria parasites in thick blood smears. This task is based on an open-source dataset\cite{35} that we term \textbf{SimuData}, where thick blood smears were stained and imaged via a modified mobile phone camera. This dataset contains 1800 large-FOV images captured at 100$\times$ magnification, each with a pixel count of 3024$\times$4032, from 150 individual patients. Each image was labelled by an expert to indicate where in the FOV the malaria parasite was visible. This task represents an ideal scenario for overlapped imaging, given the high contrast and sparsity of the parasites. 
The patients were split into training, validation, and test sets ($70\%$, $15\%$, and $15\%$ of patients, respectively). For the training and validations datasets, square regions of 96$\times$96 pixels were extracted from the full FOVs (1400 for training, 600 for validations) and marked infected if the parasite annotation lay within the inner third (32$\times$32 pixels) of the image. The datasets were balanced, so that the infection to no-infection ratio was 1:1. 

\subsubsection{DukeData}
Using the proposed overlapped imaging system (Fig. \ref{pipline}) with physical parameters specified in Table 1, we collected images from Wright-stained human peripheral blood smear slides (Carolina) to form the \textbf{DukeData} dataset. The task of interest was to automatically identify and count WBCs from within acquired images with different amounts of overlap. For experimental data collection, we first imaged peripheral blood smear slides with the seven sub-lenses of our microscope individually, using the illumination provided by a single white LED for each captured image. This produced a set of seven non-overlapped images per specimen position. Subsequently, from the seven captured ``sub-images", we cropped and resized 500 WBCs and 500 RBCs patches to be $201\times201$ pixels in size to form a ``single-lens" dataset. This dataset was used create synthetically overlapped image data using the procedure described above for \textbf{SimuData}, with which we trained CNNs for a variable number of overlaps $n$. The simulated datasets were split into training and validation with a ratio of 7:3.

Finally, $n$ = 2-7 sub-lenses were used to simultaneously image and capture physically overlapped images to form experimental datasets. We captured 35 groups of data, where one ``group'' of data includes 7 non-overlapped, single-FOV images and 6 overlapped images with different levels overlap ($n$ = 2-7, by blocking sub-apertures). The non-overlapped, single-FOV images were used to provide accurate annotations for locations of WBCs in the corresponding overlapped images. In these 35 groups, a total of 43 WBCs were observed. A sliding window approach with a 10-pixel step size was used to split the whole image into $201\times201$ pixels pitches. Patches containing whole WBCs were labeled as positive, while patches only containing RBCs and background were labeled as negative. 
The CNNs trained on synthetically overlapped data were applied to these experimentally-overlapped data to evaluate performance. 

\begin{figure}
\centering
\includegraphics[width=13cm]{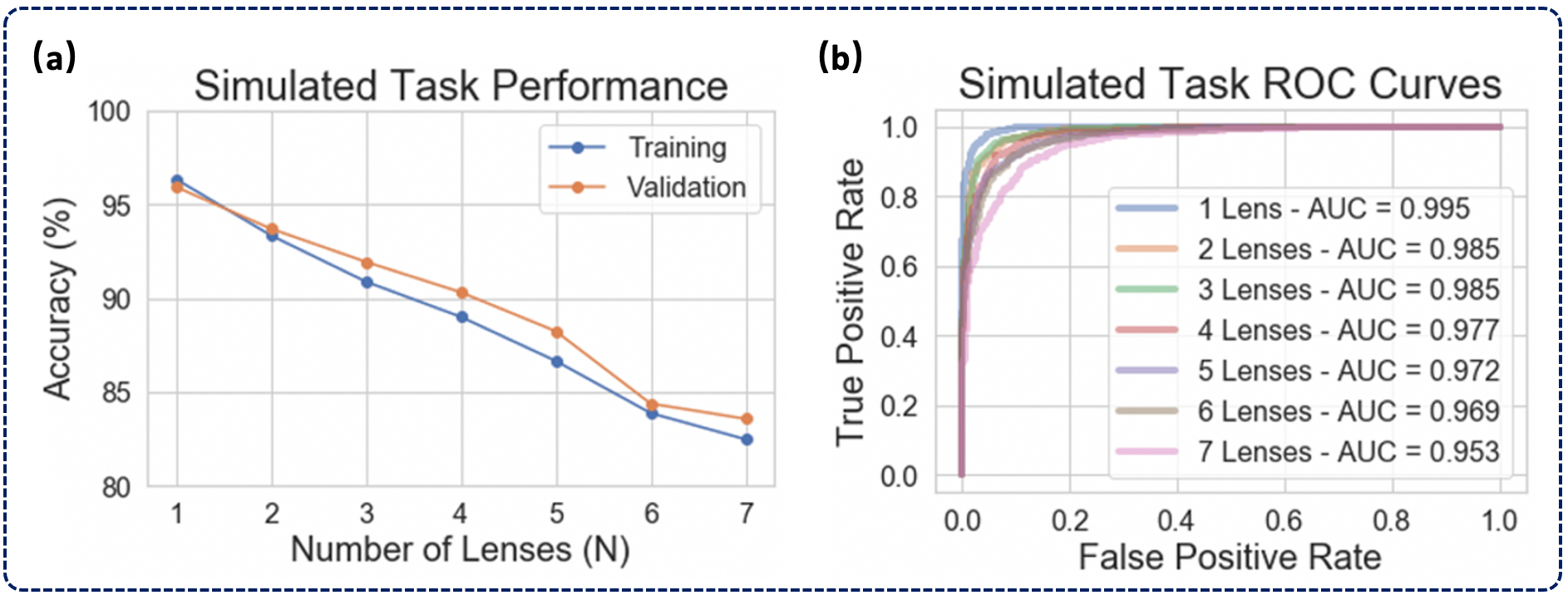}
\caption{Results of overlapped malaria parasite counting task (\textbf{SimuData}). (a) Task performance for all non-overlap and overlap conditions. (b) ROC curves for all overlap conditions.}
\label{simu_res}
\end{figure}

\section{Results}

\subsection{Simulation results}
We first investigated the impact of overlapped imaging (i.e., the resulting reduced contrast and SNR) on classification accuracy in simulation with a malaria parasite counting task (\textbf{SimuData}). We characterized classification performance across a wide range of $n$ by digitally adding images and noise as discussed above and in the Appendix (Fig. \ref{simu_res}).
 Fig. \ref{simu_res}a shows how the number of overlapped images impacts the classification task performance. In particular, we can see that although performance degrades roughly linearly with the number of overlapped images for detecting the malaria parasite, at $n$ = 7, a detection accuracy above 80\% is still maintained for both the training and validation sets.

A receiver operating characteristic (ROC) curve for classifying the malaria parasite for $n$ = 1 - 7 overlapped images is shown in Fig. \ref{simu_res}b, with the area under the curve (AUC) for each curve displayed in the legend. The ROC gives a different perspective on task performance, showing how with a relatively low false positive rate we can achieve a high true positive rate, even for the highly overlapped condition of $n$ = 7. The simulation results thus show that under a certain degree of overlap, the CNN model can still identify targets with relatively high accuracy.

\subsection{Overlapped imaging system characterization}
\begin{figure}
\centering
\includegraphics[width=13cm]{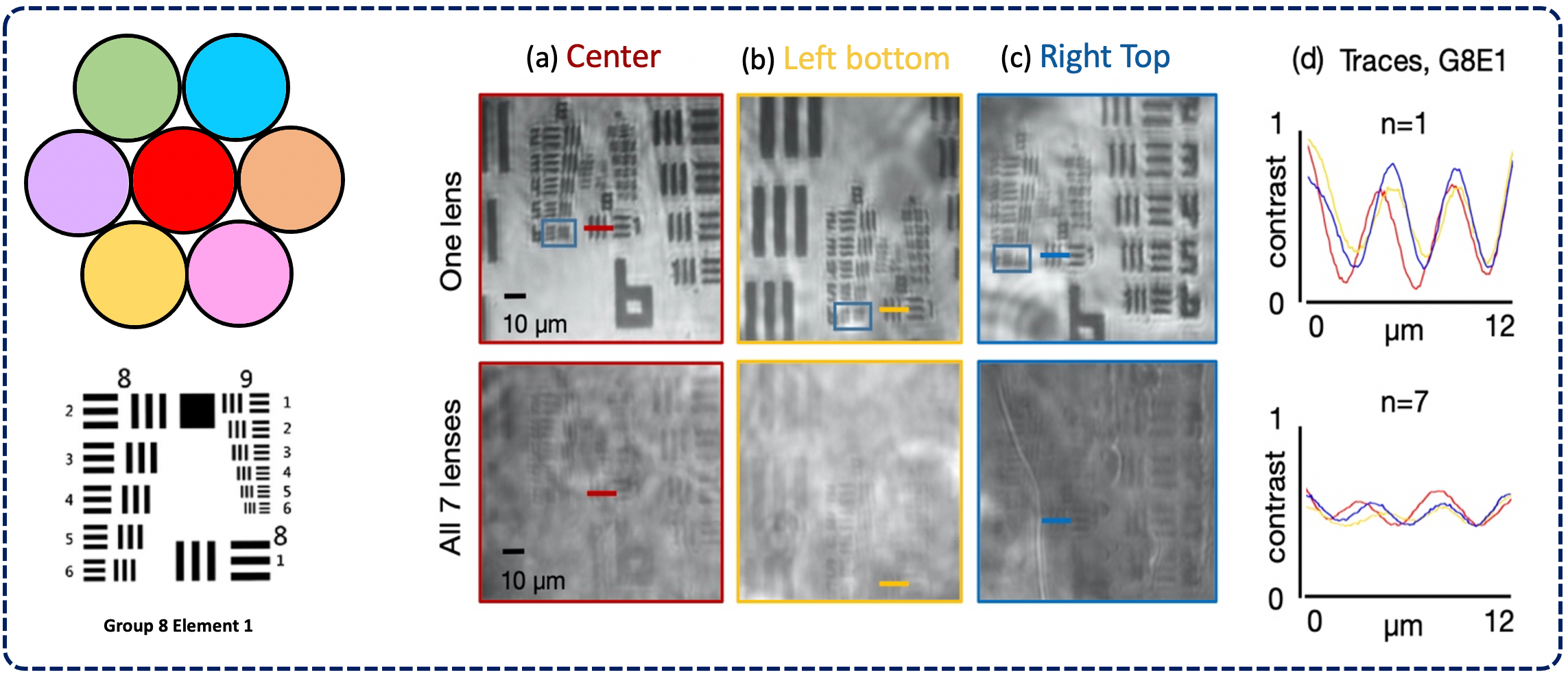}
\caption{Overlapped images of a resolution target determine system resolution and contrast. Resolution target is positioned beneath one sub-lens at a time, with all other sub-lenses blocked, for non-overlapped image capture. Example non-overlapped images (a) from the center sub-lens, (b) left-bottom sub-lens and (c) Right top sub-lens all exhibit an approximately 2-µm full-pitch resolution (see blue boxes). $n$= 7 overlapped images (bottom row), captured by illuminating all sub-FOVs at each resolution target position, demonstrate the sample is still visible. (d) Traces through group 8 element 1 (colored lines, averaged over 20 rows) show approximately a 6$\times$ higher contrast for the non-overlapped images (top) versus the corresponding overlapped images (bottom).}
\label{setup_val}
\end{figure}
We characterized the resolution of our overlapped imaging system by capturing unoverlapped images of a USAF target (Fig. \ref{setup_val}a-c). First, we used an iris to restrict incident light to illuminate the sample beneath just one sub-lens at a time. This allowed us to effectively capture each of the seven sub-images, one at a time, by simply moving the position of the iris. An example segment of one image of the resolution target, positioned and illuminated from the center sub-lens, is shown in Fig. \ref{setup_val}a. Here, we can resolve group 8 element 6, demonstrating a maximum full-pitch resolution of approximately 2 µm. We note here that the illumination source (a single white LED placed 10 mm away) provides spatially coherent illumination to the sample, suggesting that it may be possible to improve image resolution using a lower coherence source. At the same time, our source has a spatial coherence length less than w, thus ensuring that the overlapped image is an incoherent superposition of all sub-images.   


With the resolution target in the same position, we then opened the iris to illuminate the entire resolution target slide, imaging through all 7 sub-lenses and capturing an $n$ = 7 overlapped image (Fig. \ref{setup_val}a, bottom). As our USAF target slide only contained features across a 0.5 mm diameter area, most of the other sub-images that contribute to this overlapped image do not contain any obvious features and simply decrease the image contrast, as expected. We quantified the decrease in image contrast by taking traces through the resolution target at similar locations (group 8, element 1, shown as the colored horizontal line in each image). These trace values are plotted in Fig. \ref{setup_val}d (averaged over 20-pixel rows). We used the maximum difference between peak and valley in each trace curve to define the contrast in the corresponding image. Here, a normalized contrast of 0.9 for the single non-overlapped image (Fig. \ref{setup_val}d, top) dropped to approximately 0.15 for the $n$ = 7 overlapped image (Fig. \ref{setup_val}d, bottom), roughly as expected. Deviations from our expected image contrast drop of 7$\times$ may be attributed to a slightly non-uniform brightness of each sub-image across the image plane. 

\begin{figure}[H]
\centering
\includegraphics[width=13cm]{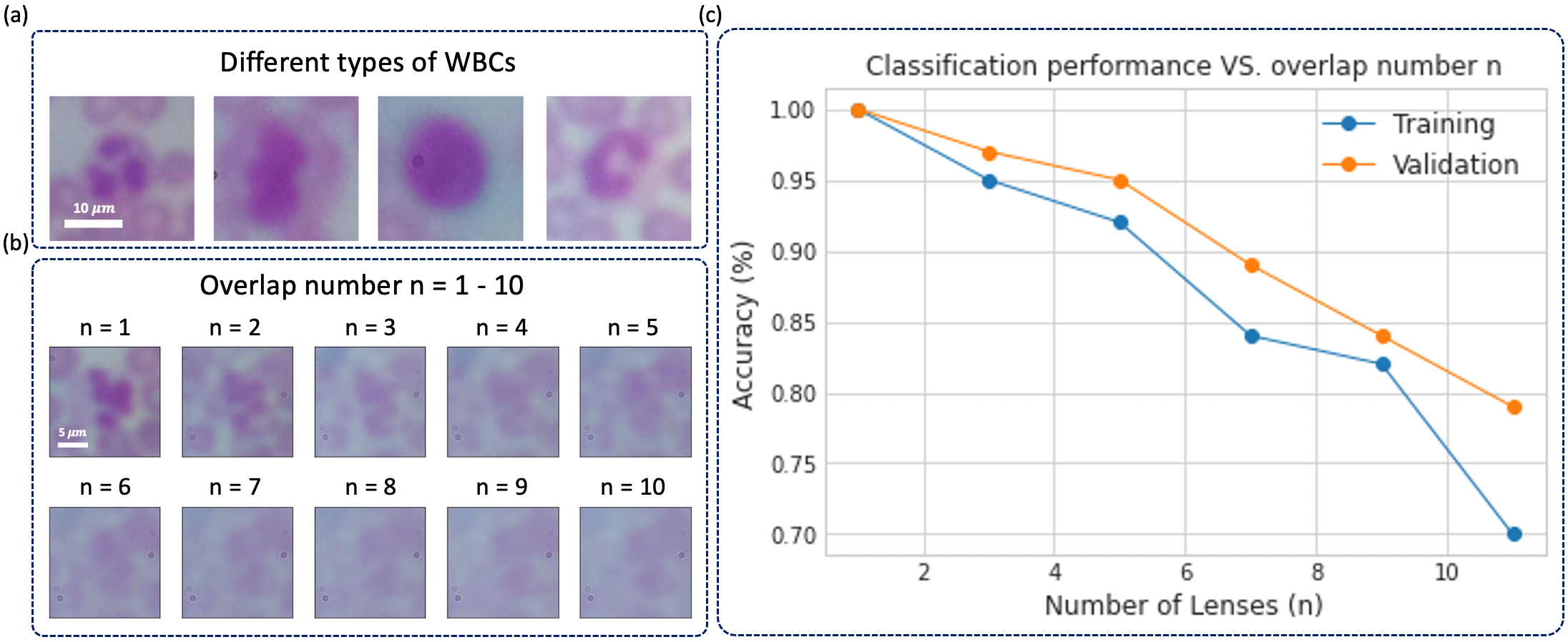}
\caption{WBC classification accuracy from digitally overlapped experimental data. (a) Different types of WBCs in \textbf{DukeData}, imaged by the central lens of our proposed microscope. 
(b) Example digitally overlapped image from experimentally acquired \textbf{DukeData}, varying from $n$ = 1 - 10. (c) Classification accuracy versus number of overlapped images.}
\label{exp_res1}
\end{figure}
We repeated this resolution target imaging experiment for all 7 of the sub-lenses, shifting both the resolution target (to lie beneath each sub-lens) and the iris (to selectively illuminate the sample) to 7 unique positions. At each position, we captured a non-overlapped image and opened the iris to capture an $n$ = 7 overlapped image. Results from performing this experiment with two other sub-lenses in our lens array are shown in Fig. \ref{setup_val}b-c. We can see that the resolution is approximately constant over the entire sub-FOV of each sub-lens, with a cutoff resolution of approximately 2 µm (see blue boxes). Furthermore, the contrast drop for each sub-lens is approximately constant at $6\times$. However, the sub-lenses do contain some non-uniform intensity variations, which we attribute to imperfections in the mounting process, as well as non-uniformities across the sample.



\subsection{Experimental results}
\begin{figure}[!t]
\centering
\includegraphics[width=10cm]{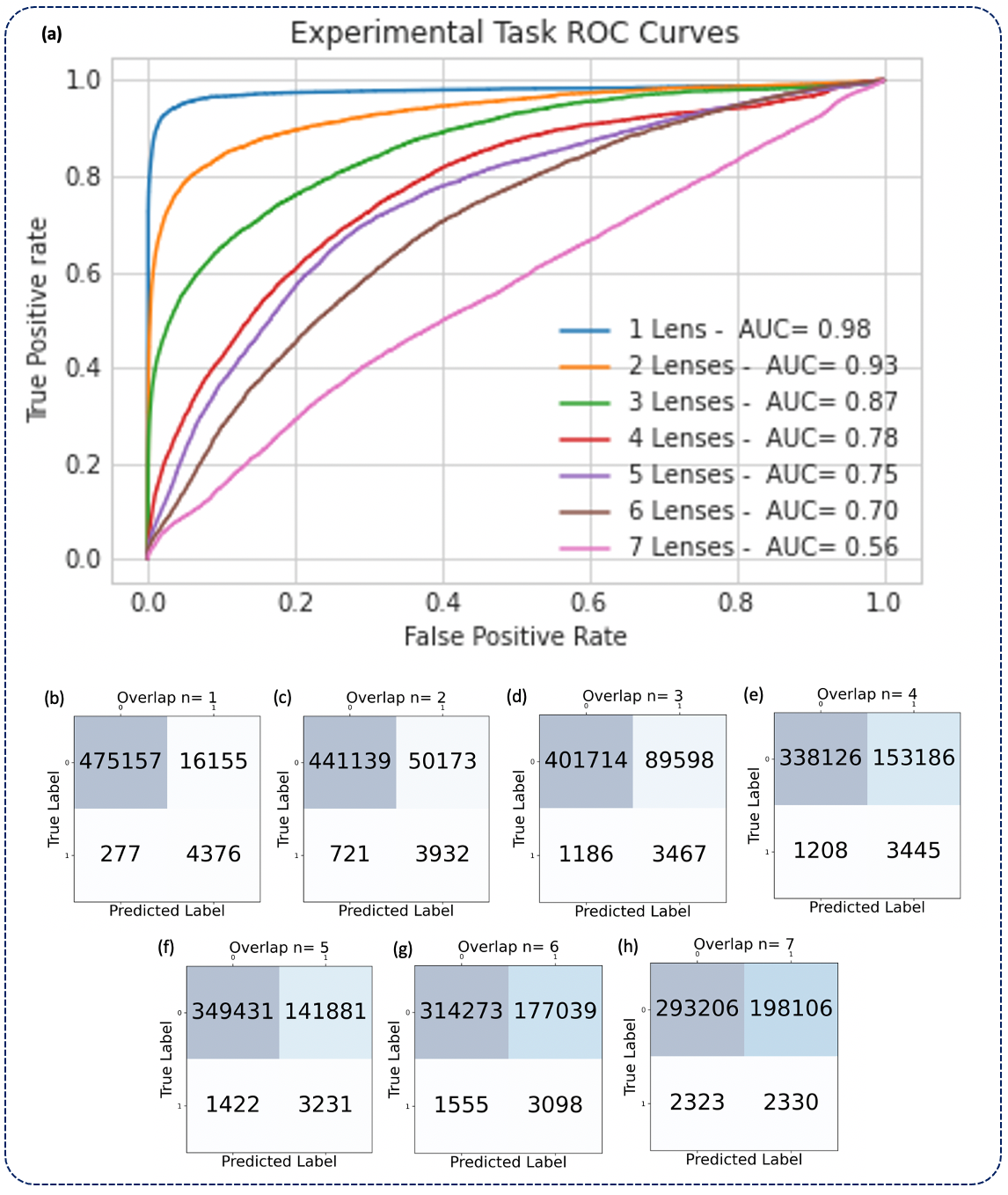}
\caption{Results for the WBC counting task using experimentally-overlapped data, reported as ROC curves (a) and confusion matrices (b-h) aggregated  under various degrees of overlap ($n$ = 1-7).}
\label{stat}
\end{figure}

\begin{figure}[!h]
\centering
\includegraphics[width=13cm]{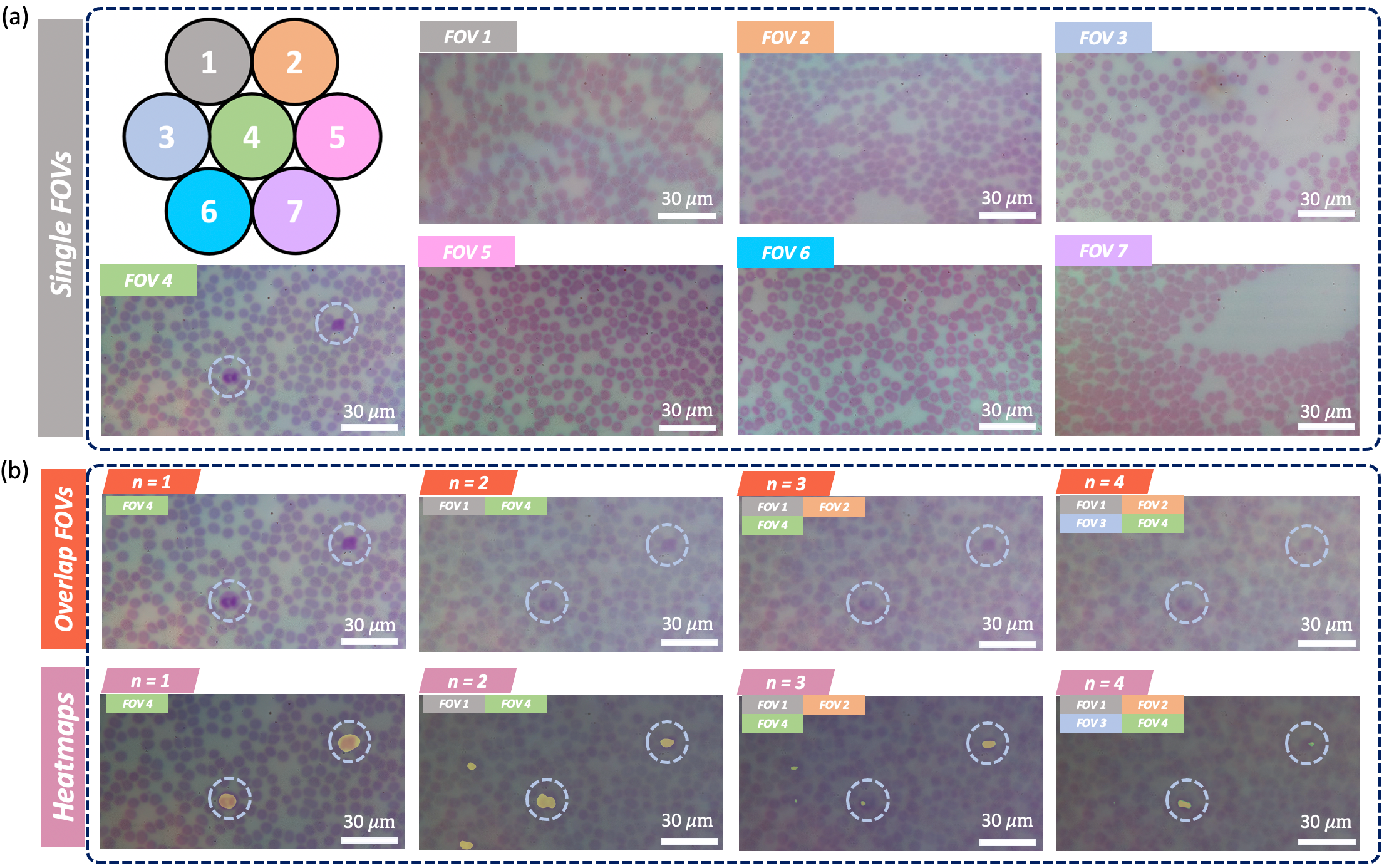}
\caption{Classification heatmaps for experimentally-overlapped images using CNNs trained on digitally-overlapped data. (a) Example single-FOV images captured by the proposed setup. (b) Overlapped images ($n$ = 1-4) captured by the proposed setup. The contributing FOVs from (a) for each overlapped image in (b) are tagged in the upper left corners. The bottom row shows the classification heatmaps generated by the CNNs overlaid on top of the overlapped images. Dotted circles identify the ground-truth locations of WBCs.}
\label{exp_res2}
\end{figure}

As a preliminary investigation of automated cell counting with our experimental microscope for overlapped imaging, we collected and processed the \textbf{DukeData} dataset (see Sec. \ref{datasets}), with results in Fig.\ref{exp_res1}-\ref{exp_res2}. When only the center lens is used, the system has a resolution comparable to that of a standard 25$\times$ microscope, allowing our proposed system to maintain crucial morphological features of different types of WBCs (see Fig. \ref{exp_res1}a). 

As a first test, we \emph{digitally-overlapped} \textbf{DukeData} images and examined WBC classification accuracy as a function of $n$ (see Fig. \ref{exp_res1}). As shown in Fig. \ref{exp_res1}b, the distinction between a WBC and red blood cells (RBC) or other background material becomes relatively visually unclear at $n=3$ or 4 overlapped images. However, our CNN classifier still maintains 95\% accuracy at $n$ = 5, and monotonically decreases with increasing overlap (Fig. \ref{exp_res1}c) as expected. Due to dynamic augmentation during training, the validation accuracies in this series of tests is slightly higher than the training accuracy. These trends are consistent with those observed in the malaria parasite counting task (Fig. \ref{simu_res}).


Next, we attempted to automatically identify WBCs in \emph{experimentally-overlapped} images, using CNN models pre-trained on \emph{digitally-overlapped} data (as described in Sec. \ref{datasets}). A unique CNN model was used for each value of $n$. Each model was independently trained 3 times with different random seeds, with their predictions averaged. Results are reported as both ROC curves (Fig. \ref{stat}a) and confusion matrices (Fig. \ref{stat}b-h) for each overlap condition ($n$ = 1-7). For the confusion matrices, we chose the threshold for the CNN outputs based on that which maximizes the geometric mean of the true positive rate and the true negative rate \cite{xie2020effect}, which accounts for dataset imbalance. In this experiment, we obtained the following aggregate detection accuracies for $n$ = 1 - 7: 96.7\%, 89.7\%, 81.7\%, 68.9\%, 71.1\%, 64.0\%, and 59.6\%.

Models obtained relatively high true positive rates for low overlap ($n$ = 1 - 4: 94.0\%, 84.5\%, 74.5\%, 74.0\%), with overall performance generally decreasing with $n$, which is consistent with our previous simulation results. The ROC curves also give similar trends (Fig. \ref{stat}a). Deviations from the expected strictly monotonic trend and the fact that performance is overall worse than with the simulated datasets are likely due to imperfections of our first experimental prototype, such as differences in brightness or color balance for different overlapping conditions.


In a third set of experiments, we applied our trained CNNs to entire images by passing a sliding window as the input to generate co-registered classification heatmaps (Fig. \ref{exp_res2}). Fig. \ref{exp_res2}a shows the non-overlapped images collected by 7 single sub-lenses of the proposed system, which were used for ground truth WBC annotations (dotted circles). Fig. \ref{exp_res2}b shows the overlapped images experimentally captured by the proposed microscope, with the classification heatmaps overlaid in the bottom row. These results qualitatively confirm that models trained with digitally overlapped data can still identify the WBCs accurately under $n$ = 4 overlap for subsequent counting, despite decreased contrast. In future work, we aim to investigate post-processing strategies to facilitate high-accuracy cell counting from such acquired overlapped image heatmaps.
\section{Discussion and conclusion}
In this work, we have demonstrated a new imaging system that can capture and overlap images of multiple independent FOVs on a common detector, which may offer a significant speed-ups for tasks requiring analysis of large FOVs. For the malaria parasite and WBC counting tasks, we investigated the relationship between CNN-based classification accuracy and number of overlapped images. For current CMOS image sensors, we showed that it is in principle possible to overlap up to 4 images while maintaining over 90\% accuracy on standard 8-bit detectors. We then presented initial results from our prototype hardware system, which can capture 2-µm resolution images over a several square millimeter areas using a set of seven small lenses. This resolution and imaging FOV of one sub-lens of our first prototype is comparable to those of a standard 25$\times$ objective lens. We also showed that CNNs trained entirely on synthetic data is able to generalize to experimental data for the WBC counting task. 

There are a number of avenues for future work that extend our proof-of-principle experiments here. Apart from exploring different CNN architectures, such as those tailored to segmentation tasks \cite{ronneberger2015u}, we could consider other types of annotation appropriate for cell-counting tasks, such as global count \cite{45}. In addition to sparse cell-counting tasks, we envision our technique being applied to other similar tasks, such as identifying defects in otherwise pristine surfaces, such as semiconductor wafers. Alternative hardware implementations may also improve performance for such tasks. For example, while our current prototype uses brightfield illumination, it may be advantageous to use darkfield illumination, which would substantially reduce the background and thus improve the contrast of the overlapped images. Further, using higher-dynamic-range sensors could compensate for lost contrast due to overlapping. Another potential direction is to use the recently proposed random access parallel microscopy setup \cite{40}, which images multiple FOVs sequentially with a single parabolic mirror as a common tube lens. Such a setup has the advantage that all sub-images overlap completely on the sensor.

We are hopeful that our initial demonstration will encourage additional exploration into the various benefits of overlapped imaging, in particular when coupled  with machine learning to automatically process the acquired data. As machine learning techniques continue to hit impressive benchmarks, we believe that our approach can leverage successive advances and inspire new research into high-throughput imaging devices that do not necessarily clearly resolve an entire scene or sample, but can still excel at specific tasks.

\bibliographystyle{unsrt}  
\bibliography{references}  

\begin{thebibliography}{10}

\bibitem{9}
Yun Liu, Krishna Gadepalli, Mohammad Norouzi, George~E Dahl, Timo Kohlberger,
  Aleksey Boyko, Subhashini Venugopalan, Aleksei Timofeev, Philip~Q Nelson,
  Greg~S Corrado, et~al.
\newblock Detecting cancer metastases on gigapixel pathology images.
\newblock {\em arXiv preprint arXiv:1703.02442}, 2017.

\bibitem{10}
Muhammad~Nasim Kashif, Shan E~Ahmed Raza, Korsuk Sirinukunwattana, Muhammmad
  Arif, and Nasir Rajpoot.
\newblock Handcrafted features with convolutional neural networks for detection
  of tumor cells in histology images.
\newblock In {\em 2016 IEEE 13th International Symposium on Biomedical Imaging
  (ISBI)}, pages 1029--1032. IEEE, 2016.

\bibitem{47}
De~Rong Loh, Wen~Xin Yong, Jullian Yapeter, Karupppasamy Subburaj, and Rajesh
  Chandramohanadas.
\newblock A deep learning approach to the screening of malaria infection:
  Automated and rapid cell counting, object detection and instance segmentation
  using mask r-cnn.
\newblock {\em Computerized Medical Imaging and Graphics}, 88:101845, 2021.

\bibitem{11}
Roarke Horstmeyer, Richard~Y Chen, Barbara Kappes, and Benjamin Judkewitz.
\newblock Convolutional neural networks that teach microscopes how to image.
\newblock {\em arXiv preprint arXiv:1709.07223}, 2017.

\bibitem{46}
Mehdi Habibzadeh, Adam Krzy{\.z}ak, and Thomas Fevens.
\newblock White blood cell differential counts using convolutional neural
  networks for low resolution images.
\newblock In {\em International Conference on Artificial Intelligence and Soft
  Computing}, pages 263--274. Springer, 2013.

\bibitem{13}
Dev~Kumar Das, R~Mukherjee, and C~Chakraborty.
\newblock Computational microscopic imaging for malaria parasite detection: a
  systematic review.
\newblock {\em Journal of microscopy}, 260(1):1--19, 2015.

\bibitem{12}
World~Health Organization.
\newblock {\em Malaria microscopy quality assurance manual-version 2}.
\newblock World Health Organization, 2016.

\bibitem{1}
Guoan Zheng, Xiaoze Ou, Roarke Horstmeyer, Jaebum Chung, and Changhuei Yang.
\newblock Fourier ptychographic microscopy: A gigapixel superscope for
  biomedicine.
\newblock {\em Optics and Photonics News}, 25(4):26--33, 2014.

\bibitem{53}
Neeta Kumar, Ruchika Gupta, and Sanjay Gupta.
\newblock Whole slide imaging (wsi) in pathology: current perspectives and
  future directions.
\newblock {\em Journal of Digital Imaging}, 33:1034--1040, 2020.

\bibitem{54}
Alexander~D Borowsky, Eric~F Glassy, William~Dean Wallace, Nathash~S
  Kallichanda, Cynthia~A Behling, Dylan~V Miller, Hemlata~N Oswal, Richard~M
  Feddersen, Omid~R Bakhtar, Arturo~E Mendoza, et~al.
\newblock Digital whole slide imaging compared with light microscopy for
  primary diagnosis in surgical pathologya multicenter, double-blinded,
  randomized study of 2045 cases.
\newblock {\em Archives of pathology \& laboratory medicine},
  144(10):1245--1253, 2020.

\bibitem{6}
Guoan Zheng, Roarke Horstmeyer, and Changhuei Yang.
\newblock Wide-field, high-resolution fourier ptychographic microscopy.
\newblock {\em Nature photonics}, 7(9):739--745, 2013.

\bibitem{konda2020fourier}
Pavan~Chandra Konda, Lars Loetgering, Kevin~C Zhou, Shiqi Xu, Andrew~R Harvey,
  and Roarke Horstmeyer.
\newblock Fourier ptychography: current applications and future promises.
\newblock {\em Optics express}, 28(7):9603--9630, 2020.

\bibitem{zheng2021concept}
Guoan Zheng, Cheng Shen, Shaowei Jiang, Pengming Song, and Changhuei Yang.
\newblock Concept, implementations and applications of fourier ptychography.
\newblock {\em Nature Reviews Physics}, 3(3):207--223, 2021.

\bibitem{brady2012multiscale}
David~J Brady, Michael~E Gehm, Ronald~A Stack, Daniel~L Marks, David~S Kittle,
  Dathon~R Golish, EM~Vera, and Steven~D Feller.
\newblock Multiscale gigapixel photography.
\newblock {\em Nature}, 486(7403):386--389, 2012.

\bibitem{fan2019video}
Jingtao Fan, Jinli Suo, Jiamin Wu, Hao Xie, Yibing Shen, Feng Chen, Guijin
  Wang, Liangcai Cao, Guofan Jin, Quansheng He, et~al.
\newblock Video-rate imaging of biological dynamics at centimetre scale and
  micrometre resolution.
\newblock {\em Nature Photonics}, 13(11):809--816, 2019.

\bibitem{bueno2014automated}
Gloria Bueno, Oscar D{\'e}niz, Mar{\'\i}a Del~Milagro Fern{\'a}ndez-Carrobles,
  Noelia V{\'a}llez, and Jes{\'u}s Salido.
\newblock An automated system for whole microscopic image acquisition and
  analysis.
\newblock {\em Microscopy research and technique}, 77(9):697--713, 2014.

\bibitem{evans2018us}
Andrew~J Evans, Thomas~W Bauer, Marilyn~M Bui, Toby~C Cornish, Helena Duncan,
  Eric~F Glassy, Jason Hipp, Robert~S McGee, Doug Murphy, Charles Myers, et~al.
\newblock Us food and drug administration approval of whole slide imaging for
  primary diagnosis: a key milestone is reached and new questions are raised.
\newblock {\em Archives of pathology \& laboratory medicine},
  142(11):1383--1387, 2018.

\bibitem{15}
Vicha Treeaporn, Amit Ashok, and Mark~A Neifeld.
\newblock Increased field of view through optical multiplexing.
\newblock {\em Optics Express}, 18(21):22432--22445, 2010.

\bibitem{16}
Ryoichi Horisaki and Jun Tanida.
\newblock Multi-channel data acquisition using multiplexed imaging with spatial
  encoding.
\newblock {\em Optics express}, 18(22):23041--23053, 2010.

\bibitem{17}
R~Hamilton Shepard, Yaron Rachlin, Vinay Shah, and Tina Shih.
\newblock Design architectures for optically multiplexed imaging.
\newblock {\em Optics express}, 23(24):31419--31435, 2015.

\bibitem{44}
Aisha Khan, Stephen Gould, and Mathieu Salzmann.
\newblock Deep convolutional neural networks for human embryonic cell counting.
\newblock In {\em European conference on computer vision}, pages 339--348.
  Springer, 2016.

\bibitem{50}
Congcong Zhang, Xiaoyan Xiao, Xiaomei Li, Ying-Jie Chen, Wu~Zhen, Jun Chang,
  Chengyun Zheng, and Zhi Liu.
\newblock White blood cell segmentation by color-space-based k-means
  clustering.
\newblock {\em Sensors}, 14(9):16128--16147, 2014.

\bibitem{49}
Syadia Nabilah~Mohd Safuan, Mohd Razali~Md Tomari, and Wan Nurshazwani~Wan
  Zakaria.
\newblock White blood cell (wbc) counting analysis in blood smear images using
  various color segmentation methods.
\newblock {\em Measurement}, 116:543--555, 2018.

\bibitem{51}
Mahdieh Poostchi, Ilker Ersoy, Katie McMenamin, Emile Gordon, Nila Palaniappan,
  Susan Pierce, Richard~J Maude, Abhisheka Bansal, Prakash Srinivasan, Louis
  Miller, et~al.
\newblock Malaria parasite detection and cell counting for human and mouse
  using thin blood smear microscopy.
\newblock {\em Journal of Medical Imaging}, 5(4):044506, 2018.

\bibitem{24}
Alex Muthumbi, Amey Chaware, Kanghyun Kim, Kevin~C Zhou, Pavan~Chandra Konda,
  Richard Chen, Benjamin Judkewitz, Andreas Erdmann, Barbara Kappes, and Roarke
  Horstmeyer.
\newblock Learned sensing: jointly optimized microscope hardware for accurate
  image classification.
\newblock {\em Biomedical optics express}, 10(12):6351--6369, 2019.

\bibitem{36}
Karen Simonyan and Andrew Zisserman.
\newblock Very deep convolutional networks for large-scale image recognition.
\newblock {\em arXiv preprint arXiv:1409.1556}, 2014.

\bibitem{he2015delving}
Kaiming He, Xiangyu Zhang, Shaoqing Ren, and Jian Sun.
\newblock Delving deep into rectifiers: Surpassing human-level performance on
  imagenet classification.
\newblock In {\em Proceedings of the IEEE international conference on computer
  vision}, pages 1026--1034, 2015.

\bibitem{74}
Zheng Ma, Lei Yu, and Antoni~B. Chan.
\newblock Small instance detection by integer programming on object density
  maps.
\newblock In {\em Proceedings of the IEEE Conference on Computer Vision and
  Pattern Recognition (CVPR)}, June 2015.

\bibitem{75}
Ann-Christin Woerl, Markus Eckstein, Josephine Geiger, Daniel~C Wagner, Tamas
  Daher, Philipp Stenzel, Aur{\'e}lie Fernandez, Arndt Hartmann, Michael Wand,
  Wilfried Roth, et~al.
\newblock Deep learning predicts molecular subtype of muscle-invasive bladder
  cancer from conventional histopathological slides.
\newblock {\em European urology}, 78(2):256--264, 2020.

\bibitem{76}
Joon-Yong Lee, Natalie~C Sadler, Robert~G Egbert, Christopher~R Anderton,
  Kirsten~S Hofmockel, Janet~K Jansson, and Hyun-Seob Song.
\newblock Deep learning predicts microbial interactions from self-organized
  spatiotemporal patterns.
\newblock {\em Computational and structural biotechnology journal},
  18:1259--1269, 2020.

\bibitem{35}
Feng Yang, Mahdieh Poostchi, Hang Yu, Zhou Zhou, Kamolrat Silamut, Jian Yu,
  Richard~J Maude, Stefan Jaeger, and Sameer Antani.
\newblock Deep learning for smartphone-based malaria parasite detection in
  thick blood smears.
\newblock {\em IEEE journal of biomedical and health informatics},
  24(5):1427--1438, 2019.

\bibitem{xie2020effect}
Chenyi Xie, Richard Du, Joshua~WK Ho, Herbert~H Pang, Keith~WH Chiu, Elaine~YP
  Lee, and Varut Vardhanabhuti.
\newblock Effect of machine learning re-sampling techniques for imbalanced
  datasets in 18 f-fdg pet-based radiomics model on prognostication performance
  in cohorts of head and neck cancer patients.
\newblock {\em European journal of nuclear medicine and molecular imaging},
  47(12):2826--2835, 2020.

\bibitem{ronneberger2015u}
Olaf Ronneberger, Philipp Fischer, and Thomas Brox.
\newblock U-net: Convolutional networks for biomedical image segmentation.
\newblock In {\em International Conference on Medical image computing and
  computer-assisted intervention}, pages 234--241. Springer, 2015.

\bibitem{45}
Yao Xue, Nilanjan Ray, Judith Hugh, and Gilbert Bigras.
\newblock Cell counting by regression using convolutional neural network.
\newblock In {\em European Conference on Computer Vision}, pages 274--290.
  Springer, 2016.

\bibitem{40}
Mishal Ashraf, Mohanan Sharika, Byu~Ri Sim, Anthony Tam, Kiamehr Rahemipour,
  Denis Brousseau, Simon Thibault, Alexander~D Corbett, and Gil Bub.
\newblock Random access parallel microscopy.
\newblock {\em Elife}, 10:e56426, 2021.

\end{thebibliography}

\appendix
\section{Proof of the noise model}

Here, we provide additional details about the noise model used to ensure our synthetically overlapped image data exhibits an experimentally accurate SNR.We prove that adding the Gaussian random variable, 
\begin{equation} \label{Z}
Z[i,j]\sim \mathcal{N}(\mu=0,\sigma^2=X_\mathit{avg}[i,j](1-1/n)2^{n_\mathit{bit}}/v),
\end{equation}
pixel-wise (indexed by $i$ and $j$) to the digitally superimposed images produces an image with the correct noise statistics in the shot noise limit. The factor of $2^{n_\mathit{bit}}/v$ in Eq. \ref{Z}, where $v$ is the pixel well depth, simply converts photoelectron count to an $n_\mathit{bit}$ value.

Let $x_q\sim \mathit{Pois}(\lambda_q)$ be the number of photons coming from the $q^\mathit{th}$ FOV, which follows a Poisson distribution with an unknown rate parameter $\lambda_q$. Then the total number of photons $x_\mathit{real}$  detected from all $n$ fields of view is $x_\mathit{real}\sim \mathit{Pois}(\Lambda)$, where $\Lambda=\sum_{q=1}^{n} 	\lambda_q$. Thus, $E(x_\mathit{real})=\mathit{Var}(x_\mathit{real})=\Lambda$, which are are the desired target statistics.
However, in the case of our digitally simulated overlapped images, we collect $n$ single field of view images with $n$ times larger illumination intensity, such that $x_q' \sim \mathit{Pois}(n\lambda_q)$. Then, as described in the main text, the simulated value is, $x_{sim}=\sum_{q=1}^{n} x_q'$, where we are ignoring discretization effects for now. Thus, $E(x_\mathit{sim})=\Lambda$, as desired, but $\mathit{Var}(x_{sim})=\Lambda/n$. We thus require a new transformed variable $x_\mathit{sim}'=f(x_\mathit{sim})$ such that $x_\mathit{sim}'$ has the correct mean and variance. We posit one possible function: 
\begin{equation}
x_\mathit{sim}'=x_\mathit{sim}+Z
\end{equation}
\begin{equation}
Z|x_{sim}\sim \mathcal{N}(\mu=0,\sigma^2=kx_\mathit{sim})
\end{equation}
where $k$ is a constant that does not depend on $\Lambda$ (as it's unknown). Because $Z$ is 0-mean, $E(x_\mathit{sim}')$ has the same (correct) mean of $\Lambda$. However, we desire a value of $k$ such that
\begin{equation} \label{var}
\mathit{Var}(x_\mathit{sim}')=\mathit{Var}(x_\mathit{sim})+\mathit{Var}(Z)+2\mathit{Cov}(x_\mathit{sim},Z)=\Lambda,
\end{equation}
as required by Poisson statistics. Of these three terms, we know only the first, $Var(x_\mathit{sim})=\Lambda/n$. The second and third terms can be computed from the joint distribution,
\begin{equation}
P(Z,x_\mathit{sim})=P(Z|x_\mathit{sim})P(x_\mathit{sim}).
\end{equation}
Unfortunately, Gaussian and Poisson distributions are not conjugate distributions, meaning further analysis would not permit analytical solutions. From Bayesian statistics, if we have a Gaussian likelihood with a known mean and unknown variance, an inverse-gamma distribution on the variance is a conjugate prior. Thus, we approximated the distribution of $x_\mathit{sim}$ with an inverse-gamma distribution that has the same mean and variance (respectively, $\Lambda$ and $\Lambda/n$):
\begin{equation}
x_\mathit{sim}\sim \mathit{InvGam}(\alpha=\Lambda n+2,\beta=\Lambda(\Lambda n+1))
\end{equation}
Then the joint distribution is a normal-inverse-gamma distribution:
\begin{equation}
Z,x_\mathit{sim}\sim \mathit{NormInvGam}(\mu=0,\lambda=1/k,
\alpha=\Lambda n+2,\beta=\Lambda(\Lambda n+1))
\end{equation}
From this joint distribution, we know that $\mathit{Var}(x_{sim})=\nicefrac{\lambda}{n}$, $\mathit{Var}(Z)=k\Lambda$ and $\mathit{Cov}(Z,x_\mathit{sim})=0$. Evaluating Eq. \ref{var}, we obtain
\begin{equation}
k=1-1/n
\end{equation}
which ensures that $\mathit{Var}(x_\mathit{sim}')=\Lambda$, thus justifying Eq. \ref{Z}.

\end{document}